\documentclass[conference]{IEEEtran}
\IEEEoverridecommandlockouts
\usepackage[caption=false,font=footnotesize]{subfig}
\usepackage{amsmath,amssymb,amsfonts,bm}
\usepackage{algorithm,algpseudocode}
\usepackage{graphicx}
\usepackage[english]{babel}
\usepackage{pgfplots}
\pgfplotsset{compat=newest}
\usepackage[square,numbers]{natbib}
\usepackage{xcolor}
\def\BibTeX{{\rm B\kern-.05em{\sc i\kern-.025em b}\kern-.08em
    T\kern-.1667em\lower.7ex\hbox{E}\kern-.125emX}}
\begin{document}

\title{DepthwiseGANs: Fast Training Generative Adversarial Networks for Realistic Image Synthesis\\
}

\author{\IEEEauthorblockN{ Mkhuseli Ngxande}
\IEEEauthorblockA{\textit{CSIR Defence, Peace, Safety and } \\
\textit{Security} \\
\textit{Optronic Sensor Systems, }\\
Pretoria, South Africa \\
Email: mngxande@csir.co.za}
\and
\IEEEauthorblockN{Jules-Raymond Tapamo }
\IEEEauthorblockA{\textit{School of Computer Engineering } \\
\textit{University of Kwa-Zulu Natal,}\\
Durban, South Africa  \\
Email: tapamoj@ukzn.ac.za}
\and
\IEEEauthorblockN{Michael Burke}
\IEEEauthorblockA{\textit{Mobile Intelligent Autonomous Systems} \\
\textit{Modelling and Digital Science}\\
\textit{Council for Scientific and Industrial Research}\\
Pretoria, South Africa\\
Email: michaelburke@ieee.org}

}
\maketitle

\begin{abstract}
Recent work has shown significant progress in the direction of synthetic data generation using Generative Adversarial Networks (GANs). GANs have been applied in many fields of computer vision including text-to-image conversion, domain transfer, super-resolution, and image-to-video applications. In computer vision, traditional GANs are based on deep convolutional neural networks. However, deep convolutional neural networks can require extensive computational resources because they are based on multiple operations performed by convolutional layers, which can consist of millions of trainable parameters. Training a GAN model can be difficult and it takes a significant amount of time to reach an equilibrium point. In this paper, we investigate the use of depthwise separable convolutions to reduce training time while maintaining data generation performance. Our results show that a DepthwiseGAN architecture can generate realistic images in shorter training periods when compared to a StarGan architecture, but that model capacity still plays a significant role in generative modelling. In addition, we show that depthwise separable convolutions perform best when only applied to the generator. For quality evaluation of generated images, we use the Fr\'{e}chet Inception Distance (FID), which compares the similarity between the generated image distribution and that of the training dataset.    
\end{abstract}

\begin{IEEEkeywords}
Synthetic Data, GANs, Depthwise Separable Convolution, FID.
\end{IEEEkeywords}

\section{INTRODUCTION AND RELATED WORK}

\begin{figure}
\centering
\includegraphics [width=0.45\textwidth]{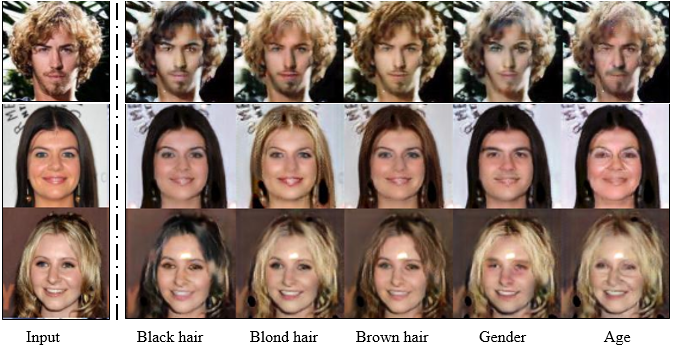}
\caption{Multi-domain image translation on the CelebA dataset using our proposed model. Given an input image from one domain, we can translate it to another, for example, changing hair color, gender or age. \label{sample}}
\end{figure}

Image-to-image translation is used in a wide variety of applications, including style transfer, super-resolution, and face synthesis \cite{zhu2017unpaired,ledig2017photo,zhang2017stackgan,huang2017beyond}. Given sufficient training data, these techniques allow one to translate an input image to a desired output, for example, changing facial attributes from a male to a female (see Figure \ref{sample}). Early work on image-to-image translation by Hertzman et al. used a non-parametric mixture model using a single input-output training image pair \cite{hertzmann2001image}. The introduction of Generative Adversarial Networks (GAN) has resulted in great success in synthesizing realistic images \cite{goodfellow2014generative,kingma2018glow,liu2017unsupervised}. Traditional GAN architectures consist of two deep convolutional neural networks (generator and discriminator), which play a mini-max game against each other. The generator produces synthetic data from a given latent sample, while the discriminator evaluates the generator's output against true data. Therefore, the generator is forced to produce appealing images to fool the discriminator. However, this competing behaviour makes it difficult to train GANs, which often require extensive computational resources \cite{chavdarova2018sgan,park2017mmgan,neyshabur2017stabilizing}. Recent research has focused on ways to stabilize the training process and using lightweight networks to speed up the process of training. The work of Arjovsky et al. proposed minimizing the Earth Mover (EM) distance in the learning distribution which helps the training process \cite{arjovsky2017wasserstein}. In addition to improving the training process, \cite{hjelm2017boundary} introduced a method that uses the estimated difference measure from the discriminator to select important weights. This provides a gradient for training the generator network and also improves the stability of the training process. Moving beyond traditional GANs, there is a wide interest in controllable GANs, for example conditional GANs, which introduce additional information to a generator to produce conditioned images \cite{zhu2017unpaired,choi2017stargan}. The StarGan architecture provides a scalable approach to perform multi-domain image-to-image translation using a single model. This is accomplished by relying on an auxiliary classifier to group labelled input data into multiple domains. This produces high quality, but relatively low resolution, synthetic images.

In this paper, we investigate the effects and benefits of substituting the traditional convolutional layers in StarGan with depthwise separable convolutions. Depthwise separable convolutions are a special case of group convolution \cite{chollet2017xception} that reduce the number of parameters in a convolution by independently applying smaller convolution filters to input tensor channels, using a 1 dimensional end stage convolution to group information from separate channels. This reduces the total number of operations required by the model. Group convolutions have already been shown to be an effective means of building lightweight models \cite{crowley2017moonshine}, but this work shows that they can affect model capacity. Our primary findings are as follows:
\begin{itemize}
\item Introducing depthwise separable convolutions in the generator reduces the number of parameters, thus resulting in faster network training.
\item The inclusion of depthwise separable convolutions in the generator is more effective then when applied in both the generator and discriminator. 
\item However, the quality of the generated images is largely dependent on network capacity, and additional layers are required to recover from capacity loss resulted from the use of depthwise separable convolutions. 
\end{itemize} 

This paper is structured as follows. Section \ref{rel_work} provides an overview of related work, which is followed by a discussion on depthwise separable convolutions. This is followed by a description of the DepthwiseGAN architecture in Section \ref{architecture} and a description of the experiments conducted in Section \ref{exps}. Finally, results and the conclusions are provided in Sections \ref{results} and \ref{concs} respectively.

\section{BACKGROUND}

\label{rel_work}

\subsection{Generative Adversarial Networks} 

GANs were first introduced by Goodfellow et al. in 2014 \cite{goodfellow2014generative}. A GAN is composed of two neural networks namely, a generator $\textit{G}$ and a discriminator $\textit{D}$, which are trained by playing a mini-max game. In the case of image translation, to learn a generator distribution $\textit{p}_{g}$ over data $x$, the generator creates a mapping function, parametrised by $\theta_g$ from a prior latent distribution $\textit{p}_{z}(z)$ to data space $G(\textit{z;}\theta_{g})$. The discriminator $D(x;\theta_{d})$, on the other hand, learns parameters $\theta_d$ to distinguish whether images are from the training data or from the generator. The mini-max game function  $V(G,D)$ is expressed as follows:

\begin{IEEEeqnarray}{lCl}
min_{G} \hspace{1mm} max_{D} \hspace{1mm} V(D,G)=\mathbb{E}_{D} + \mathbb{E}_{G} \\
\text{where } \mathbb{E}_{D}= \mathbb{E}_{\textit{x}\backsim \textit{p}_{\textbf{data}}(\textit{x}) }[ \log D(x)] \nonumber\\
\mathbb{E}_{G}= \mathbb{E}_{\textit{z}\backsim \textit{p}_{\textit{z}}(\textit{z})}[\log(1-D(G(\textit{z})))] \nonumber
\label{mini-max}
\end{IEEEeqnarray}\\ 
\subsection{Conditional GANs} 

Conditional GANs are an extension of the original GAN that includes additional information in both the generator and discriminator \cite{mirza2014conditional}. This allows for the ability to control the generated output image, so as to produce more flexible synthetic data. This information, $\textit{y}$, is typically a label applied to the resulting output image, for example, skin complexion, emotional state or hair color (see Figure \ref{sample}). The mini-max objective function from equation (\ref{mini-max}) can be expressed as follows for conditional GANs:
\begin{IEEEeqnarray}{lCl}
 min_{G}max_{D} V(D,G)=\mathbb{E}_{\textit{{x}}\backsim \textit{p}_{\textbf{data}}(\textit{{x}}) }[\log D(\textit{x$\vert$y})] + \nonumber\\ \mathbb{E}_{\textit{z}\backsim \textit{p}_{\textit{z}}}(\textit{z)}[\log(1-D(G(\textit{z$\vert$y})))]  \label{conditionnal}
\end{IEEEeqnarray}
Conditional GANs can be applied in many fields where there is limited data and data augmentation is required.

\subsection{Depthwise Separable Convolutions} 

\begin{figure}
\centering
\includegraphics [width=0.45\textwidth]{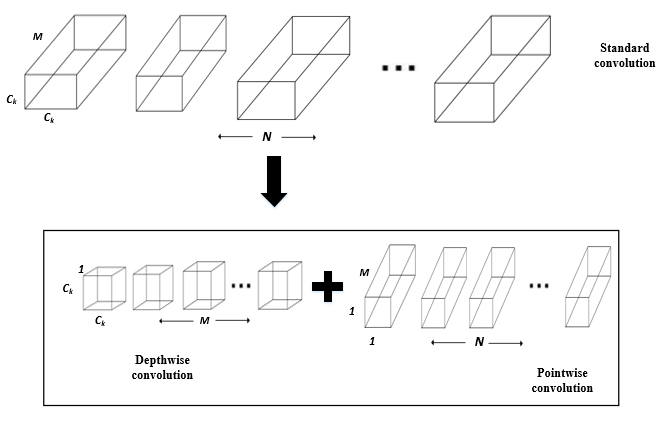}
\caption{In contrast to traditional convolution operations performed in deep learning, depthwise separable convolution perform convolutions on subsets of the input tensor and aggregate the information across these subsets using a single pointwise convolution.\label{Depth-Arch}}
\end{figure}

Depthwise separable convolutions were introduced by Chollet in 2016 as a replacement for standard Convolutions \cite{chollet2017xception}. They are a form of factorized convolutions that apply separate convolution operations to every input channel of a tensor. Depthwise separable convolutions have fewer trainable parameters, but can result in improved performance. Depthwise separable convolutions are an extension to the Xception layer \cite{szegedy2015going} and are composed of two components, namely a depthwise convolution and a pointwise convolution. The depthwise convolution performs a spatial convolution over each input channel data while the pointwise convolution layer performs a 1 $\times$ 1 convolution that merges information from the previous step across all channels. The original convolution layer dimensions can be expressed as $(H,W,C_{in},C_{out},K) $ where $ H$ is the height of an input tensor, $W$ its the width, $C$ its the depth and $K$ the size of the convolution kernel. A depthwise separable convolution operation can be expressed using dimensions $(H,W,C_{in},K)$ for the depthwise convolution and $ (H,W,C_{out})$ for the pointwise convolution. This results in an operation reduction of $\frac{1}{C_{out}} + \frac{1}{K^{2}}$. Figure \ref{Depth-Arch} illustrates the comparison between the standard convolution and depthwise separable convolution architectures.

Depthwise separable convolution have found success in a variety of applications \cite{howard2017mobilenets,Kaiser2017,Wang2018,Mahdianpari2018}. The work of Nguyen and Ray proposed an adaptive convolution block method that learns the upsampling algorithm \cite{nguyen2018generative}. They replace traditional convolutions with depthwise separable convolutions in the generator to improve the performance of a weak baseline model. Moreover, Wojna et al. have also applied depthwise separable convolutions in GAN architectures, where they introduced a bilinear additive upsampling layer, which improves performance \cite{wojna2017devil}. However, our work investigates the performance of depthwise separable convolutions on the upsampling, downsampling and bottleneck layers. This is only applied in the generator. Our results show that depthwise separable convolutions are more effective when only used on the generator, that they reduce the number of parameters, which speeds up training time, but that deeper models are required to obtain similar performance to standard convolutions. 
  
\section{DEPTHWISEGAN ARCHITECTURES}

\label{architecture}

This section describes the proposed modifications made to StarGan in order to produce lightweight networks with fewer trainable parameters on the generator network for experimental testing.

\begin{figure}
\centering
\includegraphics [width=0.45\textwidth]{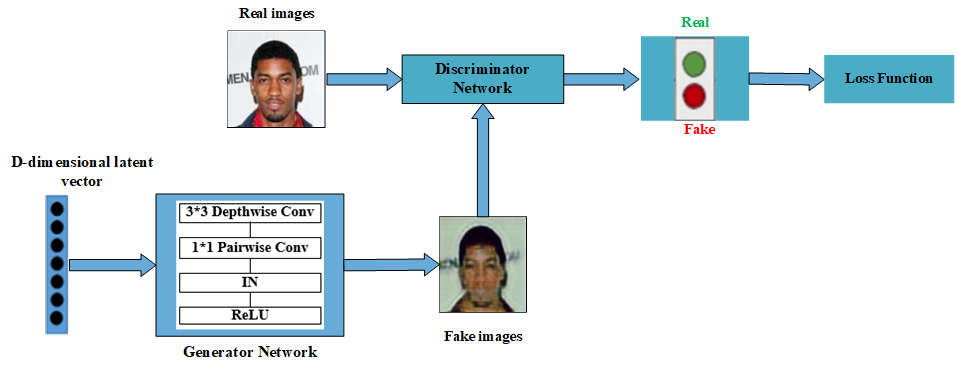}
\caption{The DepthwiseGan architecture is composed of two networks (Generator (\textit{G}) and Discriminator (\textit{D})). We used depthwise separable convolutions in the generator coupled with Instance Normalization (IN) and ReLu activation functions.\label{Arch}}
\end{figure}

The architecture used for testing is adapted from \cite{choi2017stargan}, and consists of two networks, where the generator consists of stride size of two for downsampling and 11 depthwise separable convolutions (see figure \ref{Arch}).Instance normalization is an operation of removing instance-specific contrast information from the content image which prevents mean and covariance shift and simplifies learning process \cite{ulyanov1607instance}. We applied instance normalization and ReLu activations after each pointwise convolution. For the discriminator network, a Markovian discriminator (PatchGan) \cite{Li2016} was adopted because it is a fixed-size patch discriminator that is easily applied to 256$\times $256 images. We prepared three models for training, which we term DepthwiseDG, DepthwiseG, and DeeperDepthwiseG models. The DepthwiseDG model replaces convolution layers with depthwise convolutions in both the generator and discriminator, the DepthwiseG model replaces convolutions in only the generator, while the DeeperDepthwiseG model is a deeper model with 11 depthwise separable convolution layers in the generator. All three models were compared against the original StarGan architecture.

\section{EXPERIMENTAL METHODS}
\label{exps} 

The models above were tested on a number of datasets. These are briefly described, in addition to a description of the DepthwiseGan training process. Finally, the Fr\'{e}chet Inception Distance (FID), which aims to measure the quality of generated image distribution is described.

\subsection{Datasets} 

Conditional generation of facial images was explored using three datasets. Each of these is discussed below.

\begin{LaTeXdescription}
\item[CelebA] The CelebFace Attributes dataset contains about 200k faces of celebrities \cite{Liu2015}. Each image is annotated with 40 attributes and faces cover a relatively large pose variation, with background clutter included. Images were cropped, with the faces centered and resized to $128 \times 128$ pixels. We used the seven facial domains investigated in \cite{choi2017stargan} to train our models.
\item[Stirling 3D Face Database] The Stirling 3D Face dataset consist of 3,339 images collected from 99 participants (45 males and 54 females) \cite{Stirling}. Each participant makes seven facial expressions and images are captured at four different angles. The images were cropped to 256$\times$256 pixels, with the faces are centred and then resized to 128$\times$128.
\item[RaFD] The Radboud Face Database (RaFD) consists of 49 people divided into two subsets, comprising 39 Caucasian Dutch adults (19 female and 20 male) and 10 Caucasian Dutch children (6 female and 4 male) \cite{Langner2010}. The images include eight facial expressions captured from three different angles. Images were prepared in the same way as the Stirling 3D Face Database.
\end{LaTeXdescription}

\subsection{Training process}

The DepthwiseGAN training process is illustrated in algorithm \ref{DepthwiseGAN}, and uses using the following adversarial loss \cite{choi2017stargan}:
\begin{IEEEeqnarray}{lCl}
\mathcal{L}_{adv} =\mathbb{E}_{\textit{x}}[\log D_{\textit{src}}(\textit{x})] + \nonumber\\ \mathbb{E}_{\textit{x,c}}[\log(1-D_{\textit{src}}(G(\textit{x,c})))],
\end{IEEEeqnarray}
where $G$ generates an image $ G(x,c)$ which is conditioned on both the input image $ x$ and the target domain label $c$ and $ D_{src}(x)$ is a probability distribution over sources given by $D$.
The modified objective function in \cite{choi2017stargan} is also adapted and used to optimize both the generator $G$ and discriminator $D$ as follows:
\begin{IEEEeqnarray}{lCl}
\mathcal{L}_{D} = - \mathcal{L}_{adv} + \lambda_{cls}\mathcal{L}^{\textit{r}}_{cls}, \label{full-D}
\end{IEEEeqnarray}

\begin{IEEEeqnarray}{lCl}
   \mathcal{L}_{G} = \mathcal{L}_{adv} + \lambda_{cls}\mathcal{L}^{\textit{f}}_{cls} + \lambda_{rec}\mathcal{L}_{rec},
 \label{full-G}
\end{IEEEeqnarray}
where $\lambda_{\textit{cls}} $ and $\lambda_{\textit{rec}}$ are hyper-parameters that trade-off domain classification and reconstruction losses.

\begin{algorithm}
\caption{\textbf{: Model Training}}
\label{DepthwiseGAN}
 \textbf{Input:} Given a set of real images $  \lbrace x_{1},x_{2},.......\rbrace\backsim \textit{p(\textit{x})}$, $ \lambda_{\textit{cls}}= 1 , \lambda_{\textit{rec}} =10$
\begin{algorithmic}[1]
\For{ number of training iterations }
\State Sample mini-batch of latent sample $ \lbrace \textit{z} \rbrace$ from latent prior $ \textit{p}_{\textit{z}}(\textit{z})$
\State Randomly sample mini-batch of one sample $ \lbrace \textit{x}_{1}, \textit{x}_{2},\textit{y} \rbrace$ from training set.
\State Update the parameter $ \theta_{d}$ by  descending its stochastic gradient with the Adam optimizer:
\State $ \mathcal{L}_{D} = - \mathcal{L}_{adv} + \lambda_{cls}\mathcal{L}^{\textit{r}}_{cls}  $
\State Update the generator $ \theta_{g} $ by ascending its stochastic gradient with Adam optimizer :
\State $ \mathcal{L}_{G} = \mathcal{L}_{adv} + \lambda_{cls}\mathcal{L}^{\textit{f}}_{cls} + \lambda_{rec}\mathcal{L}_{rec}$
\EndFor
\State \textbf{Output} Trained generator $ G$ and discriminator $ D$. 
\end{algorithmic}
\end{algorithm}

\begin{figure}
\centering
\includegraphics [width=0.45\textwidth]{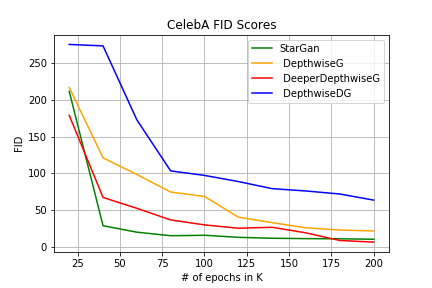}
\caption{Scores for FID on four models that we experimented on. The deeper DepthwiseGan performs similarly to the baseline model. Including depthwise convolutions in both the generator and the discriminator does not perform well.\label{fig1}}
\end{figure}
All the parameters for the training followed the StarGan training procedure and all training was conducted on a single NVIDIA Tesla K20c GPU. The Adam optimizer was used to train all the models.

\subsection{Fr\'{e}chet Inception Distance} 

Images generated using GANS can be very realistic when viewed qualitatively, but metrics are required to evaluate performance quantitatively. Here, we rely on the Fr\'{e}chet Inception Distance (FID)\cite{heusel2017gans}, which approximates the images distributions as multivariate Gaussians over pre-trained bottleneck features. The Fr\'{e}chet distance is measured between two multivariate Gaussians as follows:

\begin{IEEEeqnarray}{lCl}
FID =\parallel \mu_{\textit{r}} - \mu_{\textit{g}} \parallel^{2}  + Tr(\Sigma_{\textit{r}} + \Sigma_{\textit{g}}-2(\Sigma_{\textit{r}}\Sigma_{\textit{g}})^{1/2}).
\label{FID}
\end{IEEEeqnarray}
In this work, $\textit{X}_{r} \backsim  (\mu_{\textit{r}},\sum_{\textit{r}})$ and $\textit{X}_{\textit{g}} \backsim ( \mu_{\textit{g}},\Sigma_{\textit{g}})$ are the mean and covariance of 2048-dimensional activations extracted from a pre-trained model (Inception-v3), computed using both the training data images and samples from the generator. We resized our images to $128 \times 128$ dimensions in training phases. We sampled 140 images after every 20 000 epochs for FID evaluation.

\begin{figure}
\centering
\includegraphics [width=0.45\textwidth]{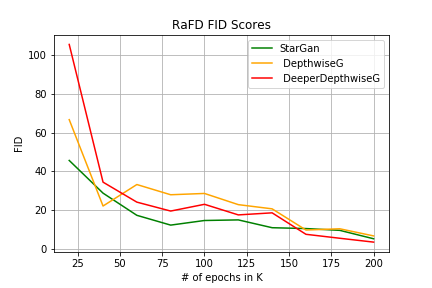}
\caption{FID scores computed for RaFD dataset. Deeper depthwiseGans obtain similar performance to starGans, and all models exhibit similar convergence after 200 epochs.\label{RafD}}
\end{figure}

\begin{figure}
\centering
\includegraphics [width=0.45\textwidth]{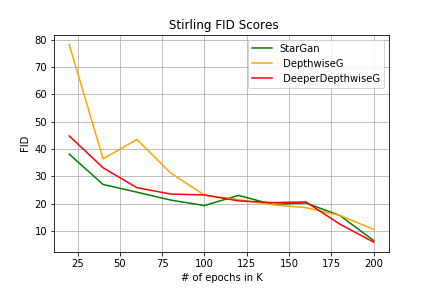}
\caption{The FID scores for the Stirling dataset exhibit similar performance to that obtained on other sets.\label{Still}}
\end{figure}

\begin{figure}
\centering
\includegraphics [width=0.45\textwidth]{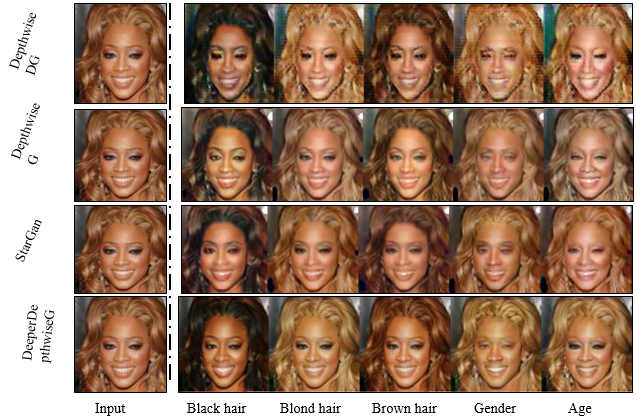}
\caption{Applying StarGan and our three models for image-to-multi domain translation shows that having depthwise separable convolutions on the discriminator can have a bad effect on the results shown on top. While adding more depthwise separable convolutions on the generator results into realistic images are shown at the bottom of the figure.\label{compareCeleb}}
\end{figure}

\section{RESULTS}
\label{results}

This section provides the experimental results obtained where we compared our models with the StarGan model in a multi-domain translation task. We used three datasets for our experiments and evaluated the quality of our results using the FID. Training the standard StarGan takes about three days (8.5M parameters for the generator and 44M parameters for discriminator). In contrast, training DepthwiseGANs is faster, the DepthwiseG model takes one day and 12 hours of training on a model with (1.5M parameters for generator and 44M parameters for discriminator), training takes two days for the DeeperDepthwiseG model (5.6M parameters for generator and 44M parameters for discriminator), and 9 hours for the DepthwiseDG  model (1.5M parameters for generator and 32M parameters for discriminator).

\subsection{Results on CelebA}

We compared DepthwiseGANs against StarGan model. Figure \ref{compareCeleb} shows the results of facial attribute transfer on the CelebA dataset using the four models. Results indicate that including depthwise convolutions in the discriminator leads to poor quality image generation. In contrast, restricting depthwise separable convolutions to the generator and increasing the depth yields results indistinguishable from StarGan. The figure shows that using depthwise separable convolutions in both networks degraded images, which appear almost cartoon-like. As a result, it appears that additional learning parameters are required to distinguish between image domains in the discriminator.

The FID measure in Figure \ref{fig1} shows that the convergence of the distribution of images generated by models using depthwise separable convolutions to the training dataset distribution is faster for StarGans, although similar performance is obtained after about 175 epochs. It is clear that deeper models perform better, indicating that the model capacity introduced by additional parameters is important if generated images are to follow a similar distribution to that of the training set.

\subsection{Results on RaFD and Stirling}

\begin{figure}
\centering
\includegraphics [width=0.45\textwidth]{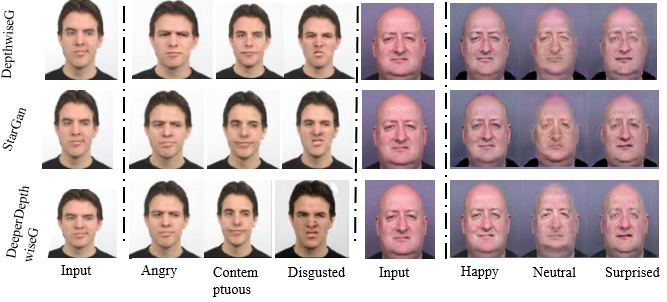}
\caption{Samples generated show that generative models with additional capacity can perform well. The RaFD and Stirling datasets are similar, but captured under different lighting conditions.\label{STFD}}
\end{figure}

We also trained our models on RaRD and Stirling datasets for facial expression transfer. Both the Stirling and RaFD datasets comprise of similar facial expressions, captured in different environments.
Results illustrated in Figure \ref{STFD} shows that the DeeperDepthwiseG model generates high-quality images when compared to the original StarGan. The FID scores in Figures \ref{RafD} and \ref{Still} show that the difference between generated and training image distributions is similar for all models. Qualitatively, images produced by the model with fewer parameters have blurry regions, and it appears that the model has limited learning capabilities due to the reduced parameter set. However, the model with more parameters generates better samples, highlighting the importance of model capacity further. 

\section{CONCLUSION}

\label{concs}

In this paper, we investigated the effects of introducing depthwise separable convolutions to the StarGan architecture. Results showed that the generation of high-quality images relies on the depth or capacity of the network, and that GANs trained using depthwise separable convolutions require additional capacity to produce similar results to StarGans, although they do reduce computational requirements substantially. 

\bibliographystyle{IEEEtran}  
\bibliography{refs}

\begin{thebibliography}{10}
\providecommand{\url}[1]{#1}
\csname url@samestyle\endcsname
\providecommand{\newblock}{\relax}
\providecommand{\bibinfo}[2]{#2}
\providecommand{\BIBentrySTDinterwordspacing}{\spaceskip=0pt\relax}
\providecommand{\BIBentryALTinterwordstretchfactor}{4}
\providecommand{\BIBentryALTinterwordspacing}{\spaceskip=\fontdimen2\font plus
\BIBentryALTinterwordstretchfactor\fontdimen3\font minus
  \fontdimen4\font\relax}
\providecommand{\BIBforeignlanguage}[2]{{%
\expandafter\ifx\csname l@#1\endcsname\relax
\typeout{** WARNING: IEEEtran.bst: No hyphenation pattern has been}%
\typeout{** loaded for the language `#1'. Using the pattern for}%
\typeout{** the default language instead.}%
\else
\language=\csname l@#1\endcsname
\fi
#2}}
\providecommand{\BIBdecl}{\relax}
\BIBdecl

\bibitem{zhu2017unpaired}
J.-Y. Zhu, T.~Park, P.~Isola, and A.~A. Efros, ``Unpaired image-to-image
  translation using cycle-consistent adversarial networks,'' \emph{arXiv
  preprint}, 2017.

\bibitem{ledig2017photo}
C.~Ledig, L.~Theis, F.~Husz{\'a}r, J.~Caballero, A.~Cunningham, A.~Acosta,
  A.~P. Aitken, A.~Tejani, J.~Totz, Z.~Wang \emph{et~al.}, ``Photo-realistic
  single image super-resolution using a generative adversarial network.'' in
  \emph{CVPR}, vol.~2, no.~3, 2017, p.~4.

\bibitem{zhang2017stackgan}
H.~Zhang, T.~Xu, H.~Li, S.~Zhang, X.~Huang, X.~Wang, and D.~Metaxas,
  ``Stackgan: Text to photo-realistic image synthesis with stacked generative
  adversarial networks,'' \emph{arXiv preprint}, 2017.

\bibitem{huang2017beyond}
R.~Huang, S.~Zhang, T.~Li, R.~He \emph{et~al.}, ``Beyond face rotation: Global
  and local perception gan for photorealistic and identity preserving frontal
  view synthesis,'' \emph{arXiv preprint arXiv:1704.04086}, 2017.

\bibitem{hertzmann2001image}
A.~Hertzmann, C.~E. Jacobs, N.~Oliver, B.~Curless, and D.~H. Salesin, ``Image
  analogies,'' in \emph{Proceedings of the 28th annual conference on Computer
  graphics and interactive techniques}.\hskip 1em plus 0.5em minus 0.4em\relax
  ACM, 2001, pp. 327--340.

\bibitem{goodfellow2014generative}
I.~Goodfellow, J.~Pouget-Abadie, M.~Mirza, B.~Xu, D.~Warde-Farley, S.~Ozair,
  A.~Courville, and Y.~Bengio, ``Generative adversarial nets,'' in
  \emph{Advances in neural information processing systems}, 2014, pp.
  2672--2680.

\bibitem{kingma2018glow}
D.~P. Kingma and P.~Dhariwal, ``Glow: Generative flow with invertible 1x1
  convolutions,'' \emph{arXiv preprint arXiv:1807.03039}, 2018.

\bibitem{liu2017unsupervised}
M.-Y. Liu, T.~Breuel, and J.~Kautz, ``Unsupervised image-to-image translation
  networks,'' in \emph{Advances in Neural Information Processing Systems},
  2017, pp. 700--708.

\bibitem{chavdarova2018sgan}
T.~Chavdarova and F.~Fleuret, ``Sgan: An alternative training of generative
  adversarial networks,'' in \emph{Proceedings of the IEEE international
  conference on Computer Vision and Pattern Recognition}, 2018.

\bibitem{park2017mmgan}
N.~Park, A.~Anand, J.~R.~A. Moniz, K.~Lee, T.~Chakraborty, J.~Choo, H.~Park,
  and Y.~Kim, ``Mmgan: Manifold matching generative adversarial network for
  generating images,'' \emph{arXiv preprint arXiv:1707.08273}, 2017.

\bibitem{neyshabur2017stabilizing}
B.~Neyshabur, S.~Bhojanapalli, and A.~Chakrabarti, ``Stabilizing gan training
  with multiple random projections,'' \emph{arXiv preprint arXiv:1705.07831},
  2017.

\bibitem{arjovsky2017wasserstein}
M.~Arjovsky, S.~Chintala, and L.~Bottou, ``Wasserstein gan,'' \emph{arXiv
  preprint arXiv:1701.07875}, 2017.

\bibitem{hjelm2017boundary}
R.~D. Hjelm, A.~P. Jacob, T.~Che, A.~Trischler, K.~Cho, and Y.~Bengio,
  ``Boundary-seeking generative adversarial networks,'' \emph{arXiv preprint
  arXiv:1702.08431}, 2017.

\bibitem{choi2017stargan}
Y.~Choi, M.~Choi, M.~Kim, J.-W. Ha, S.~Kim, and J.~Choo, ``Stargan: Unified
  generative adversarial networks for multi-domain image-to-image
  translation,'' \emph{arXiv preprint}, vol. 1711, 2017.

\bibitem{chollet2017xception}
F.~Chollet, ``Xception: Deep learning with depthwise separable convolutions,''
  \emph{arXiv preprint}, pp. 1610--02\,357, 2017.

\bibitem{crowley2017moonshine}
E.~J. Crowley, G.~Gray, and A.~Storkey, ``Moonshine: Distilling with cheap
  convolutions,'' \emph{arXiv preprint arXiv:1711.02613}, 2017.

\bibitem{mirza2014conditional}
M.~Mirza and S.~Osindero, ``Conditional generative adversarial nets,''
  \emph{arXiv preprint arXiv:1411.1784}, 2014.

\bibitem{szegedy2015going}
C.~Szegedy, W.~Liu, Y.~Jia, P.~Sermanet, S.~Reed, D.~Anguelov, D.~Erhan,
  V.~Vanhoucke, and A.~Rabinovich, ``Going deeper with convolutions,'' in
  \emph{Proceedings of the IEEE conference on computer vision and pattern
  recognition}, 2015, pp. 1--9.

\bibitem{howard2017mobilenets}
A.~G. Howard, M.~Zhu, B.~Chen, D.~Kalenichenko, W.~Wang, T.~Weyand,
  M.~Andreetto, and H.~Adam, ``Mobilenets: Efficient convolutional neural
  networks for mobile vision applications,'' \emph{arXiv preprint
  arXiv:1704.04861}, 2017.

\bibitem{Kaiser2017}
{\L}.~Kaiser, A.~N. Gomez, and F.~Chollet, ``{Depthwise Separable Convolutions
  for Neural Machine Translation},'' 2017.

\bibitem{Wang2018}
G.~Wang, G.~Yuan, T.~Li, and M.~Lv, ``{An multi-scale learning network with
  depthwise separable convolutions},'' \emph{Transactions on Computer Vision
  and Applications}, pp. 1--8, 2018.

\bibitem{Mahdianpari2018}
M.~Mahdianpari, B.~Salehi, M.~Rezaee, F.~Mohammadimanesh, and Y.~Zhang, ``{Very
  Deep Convolutional Neural Networks for Complex Land Cover Mapping Using
  Multispectral Remote Sensing Imagery},'' \emph{MDPI}, 2018.

\bibitem{nguyen2018generative}
N.~M. Nguyen and N.~Ray, ``Generative adversarial networks using adaptive
  convolution,'' \emph{arXiv preprint arXiv:1802.02226}, 2018.

\bibitem{wojna2017devil}
Z.~Wojna, V.~Ferrari, S.~Guadarrama, N.~Silberman, L.-C. Chen, A.~Fathi, and
  J.~Uijlings, ``The devil is in the decoder,'' \emph{arXiv preprint
  arXiv:1707.05847}, 2017.

\bibitem{ulyanov1607instance}
D.~Ulyanov, A.~Vedaldi, and V.~Lempitsky, ``Instance normalization: The missing
  ingredient for fast stylization. arxiv 2016,'' \emph{arXiv preprint
  arXiv:1607.08022}.

\bibitem{Li2016}
C.~Li and M.~Wand, ``{Precomputed Real-Time Texture Synthesis with Markovian
  Generative Adversarial Networks},'' pp. 1--17, 2016.

\bibitem{Liu2015}
Z.~Liu, P.~Luo, X.~Wang, and X.~Tang, ``{Deep Learning Face Attributes in the
  Wild},'' \emph{ICCV}, pp. 3730--3738, 2015.

\bibitem{Stirling}
\BIBentryALTinterwordspacing
(2018) Psychological image collection at stirling (pics). Accessed 2018/09/10.
  [Online]. Available: \url{http://www.pics.stir.ac.uk}
\BIBentrySTDinterwordspacing

\bibitem{Langner2010}
O.~Langner, R.~Dotsch, G.~Bijlstra, D.~H.~J. Wigboldus, S.~T. Hawk, A.~Van,
  O.~Langner, R.~Dotsch, G.~Bijlstra, and D.~H.~J. Wigboldus, ``{Presentation
  and validation of the Radboud Faces Database},'' no. 907172236, 2010.

\bibitem{heusel2017gans}
M.~Heusel, H.~Ramsauer, T.~Unterthiner, B.~Nessler, and S.~Hochreiter, ``Gans
  trained by a two time-scale update rule converge to a local nash
  equilibrium,'' in \emph{Advances in Neural Information Processing Systems},
  2017, pp. 6626--6637.

\end{thebibliography}

\end{document}